\documentclass[letterpaper, 10 pt, conference]{ieeeconf}  
\pdfoutput=1

\IEEEoverridecommandlockouts                              

\usepackage{booktabs}
\usepackage{multicol}
\usepackage{paralist}

\usepackage{floatrow}
\usepackage{algpseudocode,algorithm,algorithmicx}
\algnewcommand\algorithmicforeach{\textbf{for each}}
\algdef{S}[FOR]{ForEach}[1]{\algorithmicforeach\ #1\ \algorithmicdo}

\algrenewcommand\algorithmicrequire{\textbf{Precondition:}}
\algrenewcommand\algorithmicensure{\textbf{Postcondition:}}

\usepackage{times}
\usepackage{epsfig}
\usepackage{graphicx}
\usepackage{amsmath}
\usepackage{amssymb}
\DeclareMathOperator*{\argmax}{arg\,max}

\usepackage[obeyFinal]{easy-todo}

\title{\LARGE \bf
The Importance of Metric Learning for Robotic Vision:\\Open Set Recognition and Active Learning
}

\author{Benjamin J. Meyer and Tom Drummond
\thanks{This research was supported by the Australian Research Council Centre of Excellence for Robotic Vision (project number CE140100016).}%
\thanks{The authors are with the ARC Centre of Excellence for Robotic Vision, Monash University.\newline
{\tt \{benjamin.meyer, tom.drummond\}@monash.edu }}
}

\begin{document}

\maketitle
\thispagestyle{empty}
\pagestyle{empty}

\begin{abstract}
State-of-the-art deep neural network recognition systems are designed for a static and closed world. It is usually assumed that the distribution at test time will be the same as the distribution during training. As a result, classifiers are forced to categorise observations into one out of a set of predefined semantic classes. Robotic problems are dynamic and open world; a robot will likely observe objects that are from outside of the training set distribution. Classifier outputs in robotic applications can lead to real-world robotic action and as such, a practical recognition system should not silently fail by confidently misclassifying novel observations. We show how a deep metric learning classification system can be applied to such open set recognition problems, allowing the classifier to label novel observations as unknown. Further to detecting novel examples, we propose an open set active learning approach that allows a robot to efficiently query a user about unknown observations. Our approach enables a robot to improve its understanding of the true distribution of data in the environment, from a small number of label queries. Experimental results show that our approach significantly outperforms comparable methods in both the open set recognition and active learning problems.
\end{abstract}

\section{Introduction}
\begin{figure}
\includegraphics[width=\textwidth]{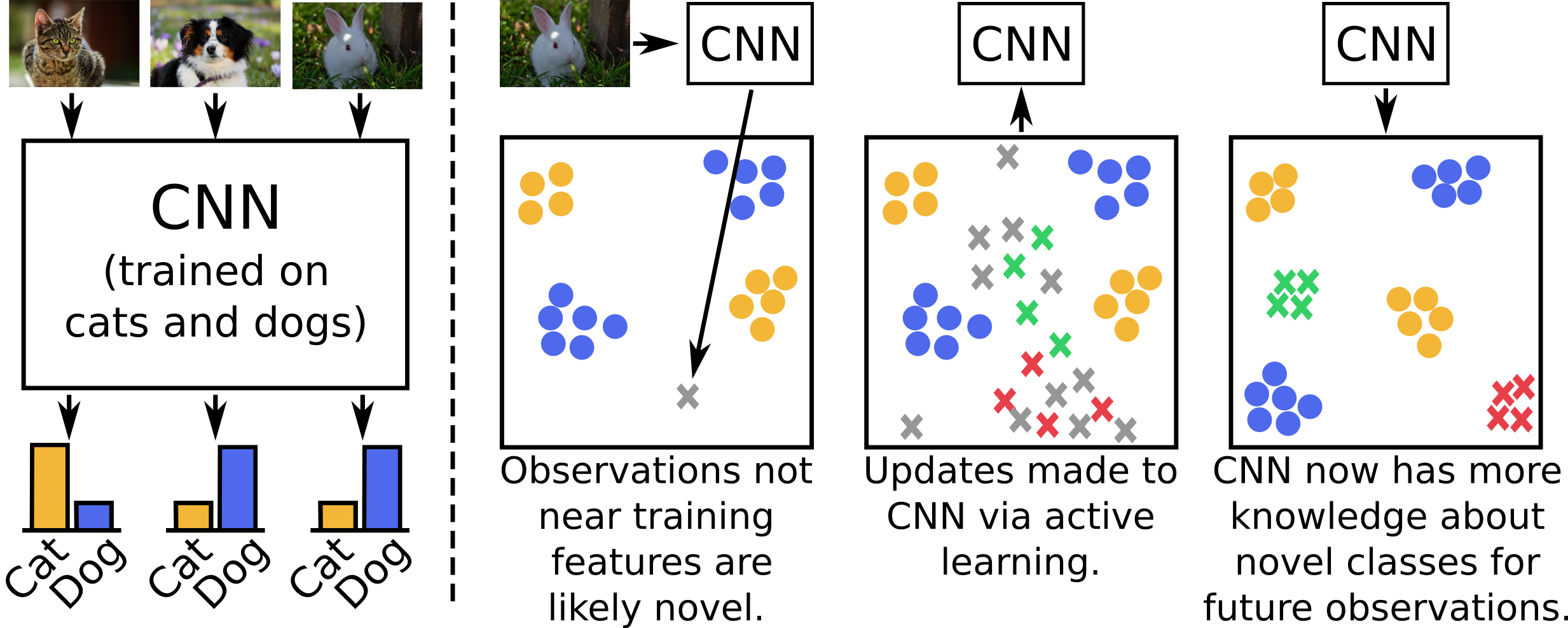}
\begin{tabular}{@{}c@{}c}
\small
\hspace*{1cm}(a) \hspace*{4.4cm}& \small(b)\hspace*{2.5cm}
\normalsize
\end{tabular}
\caption{(a) Motivation for our approach. Conventional classifiers will silently fail when observing a novel example. (b) Overview of our approach. }
\label{motivation_overview}
\end{figure}

Robotic applications demand special considerations when designing a visual recognition system. The open set nature of robotics means that a robot will encounter observations belonging to novel, out-of-distribution classes. Any classifier prediction in a robotic environment can trigger some sort of costly robotic action. As such, a recognition system must not silently fail when observing a novel example by incorrectly predicting a label from the training set distribution, as shown in Figure \ref{motivation_overview}a. Additionally, a robotic vision system should not cease learning after the initial training phase. The distribution of data in the training set will undoubtedly vary from the true distribution of data in the robot's operating environment. By sampling data from the environment and interactively querying a human user about novel observations, a robotic vision system can continue to improve its understanding of the real-world data distribution, as shown in Figure \ref{motivation_overview}b.

Detecting out-of-distribution observations is known as novelty detection and the problem of both classifying in-distribution observations with the correct class label and detecting novel examples is known as open set recognition. The task of interactively querying a user for labels is referred to as active learning. If a recognition system can select the most informative observations for labelling, the model can efficiently learn from a small number of labelled examples. This is important for robotics, as the number of observations may be very large but the labelling budget is likely to be small. In this work we focus on the active learning of novel classes, also referred to as open set active learning. A model trained on known classes is deployed in an environment containing both known and novel classes. The active learning algorithm aims to learn about the novel class distribution from as few human-labelled examples as possible.

Conventional classification approaches, such as convolutional neural networks (CNNs) with softmax classifiers \cite{NIPS2012_4824,Simonyan14c,7298594,7780459}, are closed set by design. As such, these commonly used methods are limited in their ability to detect novel examples. Softmax-based CNNs are not only forced to predict a known label for out-of-distribution examples, but often do so with high confidence \cite{DBLP:journals/corr/HendrycksG16c}. Approaches that are limited in their ability to detect novel examples, are also limited in their ability to learn from and improve their understanding of the corresponding unknown classes. As a result, these conventional softmax-based approaches are not suitable for open set robotic vision problems.

Deep metric learning algorithms learn a transformation from the image space to a feature embedding space, in which distance is a measure of semantic similarity. 
State-of-the-art deep metric learning models demonstrate an impressive aptitude for transfer learning \cite{7298682,7780803,NIPS2016_6200,7780950,DBLP:journals/corr/SongJR016,1704.01285,DBLP:journals/corr/MeyerHD17}, meaning that features are likely to be co-located based on class, even when those classes are outside of the training distribution. 
This not only allows for the reliable detection of novel examples, but also provides a meaningful way of determining an observation's informativeness of the true class distribution. This knowledge enables efficient querying in an active learning setting, allowing the model to learn about novel classes from a small amount of labelled examples.

An overview of our approach is shown in Figure \ref{motivation_overview} and the main contributions of this paper are as follows:
\begin{compactitem}
  \item We present a deep metric learning approach to novelty detection and open set recognition, showing that it outperforms conventional CNNs and purpose built novelty detectors (Sections \ref{anomaly_detection} and \ref{novelty_results}).
  \item We propose an open set active learning approach using metric spaces, which allows a model to efficiently learn about observed novel classes (Section \ref{active}).
  \item We show that our proposed approach to active learning significantly outperforms comparable methods at small labelling budgets (Section \ref{active_results}). 
  \item For a labelling budget of zero, we investigate if the representation of observed novel classes can be improved using unsupervised pseudo-labels (Section \ref{pseudo_labels_sec}). 
  
\end{compactitem}

\section{Related Work}
\subsection{Novelty Detection and Open Set Recognition} An in-depth review of classical novelty detection approaches can be found in the survey by Pimentel \textit{et al.} \cite{PIMENTEL2014215}. Common methods include \textit{probabilistic approaches} \cite{grubbs1969procedures, aggarwal2008outlier} that estimate the probability density function of the data, \textit{distance approaches} \cite{10.1007/3-540-45681-3_2, 1334558} that assume novel examples are located far from known examples, \textit{domain approaches} \cite{1007557, scholkopf2000support} that treat the task as a binary classification problem and \textit{information-theoretic approaches} \cite{10.1007/11538059_42, 4470225} that analyse the information content of data. Novelty detection is related to the task of anomaly and outlier detection \cite{chandola2009anomaly}.

Recent works that make use of deep CNNs include a Generative Adversarial Network approach \cite{DBLP:journals/corr/abs-1802-10560}, in which a multi-class discriminator is trained with a generator that creates data from both known and novel distributions. Mandelbaum and Weinshall \cite{DBLP:journals/corr/abs-1709-09844} propose a density-based confidence score that can be applied to novelty detection, as an alternative to confidence scores based on softmax probabilities \cite{DBLP:journals/corr/HendrycksG16c}. 

Bendale and Boult \cite{bendale2016towards} propose an open set version of a softmax classifier named OpenMax. Class probabilities are revised using a meta-recognition Weibull model fitted on distances between activation vectors and per-class mean activation vectors. A pseudo-class representing unknown classes is introduced, allowing direct measurement of novelty.

Liang \textit{et al.} \cite{liang2018enhance} introduce an out-of-distribution detector called ODIN that operates on pre-trained softmax-based networks. The authors use softmax temperature scaling and input pre-processing to push softmax scores from known and novel classes further apart. This method requires a forward pass, backward pass and second forward pass through the network to perform novelty detection.

A contrastive loss metric learning approach is proposed in \cite{masana2018metric}. However, the classification performance is poor, making it unsuitable for direct use in open set recognition. Further, the work proposed in \cite{masana2018metric}, along with \cite{lee2017training}, requires out-of-distribution examples during training.

\subsection{Active Learning} Classic methods of active learning include uncertainty approaches \cite{10.1007/978-1-4471-2099-5_1, 10.1007/3-540-44816-0_31, 5206627} and decision-theoretic approaches \cite{settles2008multiple, Roy:2001:TOA:645530.655646, zhu2003combining}. A comprehensive review of these methods can be found in Settles' survey \cite{settles.tr09}. Recent works have investigated active learning with CNNs \cite{7508942,stark-gcpr15,DBLP:journals/corr/ZhuB17,DBLP:journals/corr/abs-1708-02383,2017arXiv170800489S}. Approaches include framing active learning as a reinforcement learning problem \cite{DBLP:journals/corr/abs-1708-02383}, generative adversarial active learning \cite{DBLP:journals/corr/ZhuB17} and a core-set based approach \cite{2017arXiv170800489S}. These methods aim to select a subset that best represents the entire set of unlabelled examples, for the purpose of initial training. Our method is focused on learning from observed novel classes that are not present in the existing training set. This means that rather than selecting a subset that best represents the entire unlabelled set, we want to select a subset that best represents the novel classes in the unlabelled set. In other words, we don't want to waste our limited labelling budget on classes that are already well learned.

\section{Metric Learning for Open Set Problems}

We first introduce the notation used for the remainder of this paper. For a given observation with label $y$, a trained neural network produces a $\beta$-dimensional feature embedding $\mathbf{x} = [x^{(1)},\mathellipsis , x^{(\beta)}]^T$ in a metric space $\mathcal{M}$. The set of $\alpha$ labelled training feature embeddings is denoted as $C = [\mathbf{c}_1,\mathellipsis ,\mathbf{c}_\alpha]$, with $\mathbf{c}_i = [c_i^{(1)},\mathellipsis , c_i^{(\beta)}]^T$ corresponding to the feature embedding vector of the $i$-th labelled training image.

\subsection{Deep Metric Learning} \label{dml_section}
Deep metric learning refers to metric learning approaches that make use of deep convolutional neural networks (CNNs). Unlike conventional classification models, such as a CNN with a softmax classifier, metric learning algorithms aim to learn a transformation from the image space to a feature embedding space $\mathcal{M}$, in which distance is a measure of semantic similarity.
Deep metric learning approaches learn feature embeddings that are amenable to transfer learning \cite{7298682,7780803,NIPS2016_6200,7780950,DBLP:journals/corr/SongJR016,1704.01285,DBLP:journals/corr/MeyerHD17}. This suggests that such models are suitable for detecting novel examples.

The deep metric learning approach we use in this paper is described in \cite{DBLP:journals/corr/MeyerHD17}. 
A Gaussian kernel, or radial basis function, is centred on each training feature embedding.
The probability that example $\mathbf{x}$ has class label $l$ is computed as:
\begin{equation} \label{gaussian_sum}
    Pr(y = l \vert \mathbf{x}) = \frac{\sum_{j \in S,j \in l} \exp{\left(\frac{-\lvert \mathbf{x} - \mathbf{c}_j \rvert^2}{2\sigma^2} \right)}}{\sum_{k \in S} \exp{\left(\frac{-\lvert \mathbf{x} - \mathbf{c}_k \rvert^2}{2\sigma^2} \right)}},
\end{equation}
where $\sigma$ is a shared standard deviation and $S \subseteq C$ is the set of training nearest neighbours for example $\mathbf{x}$.

During training the Gaussian kernels pull examples of the same class together and push examples of different classes apart. The loss for a given training example is the negative logarithm of the true class probability. The approach is made scalable to large numbers of classes and examples through the use of fast approximate nearest neighbour search. Training is made feasible and efficient by periodic asynchronous updates of the training embeddings and nearest neighbours, negating the need to do so after every network update.

Although many deep metric learning approaches perform well on transfer learning tasks, the feature embeddings learned by commonly used triplet approaches are not well suited to classification \cite{rippel2016metric}. In contrast, the Gaussian kernel approach performs well for transfer learning as well as classification, outperforming softmax classification on several datasets \cite{DBLP:journals/corr/MeyerHD17}. This makes the model suitable for our open set recognition and active learning setting.

\subsection{Novelty Detection and Open Set Recognition} \label{anomaly_detection}
We now describe the problem of detecting out-of-distribution observations. Let $\mathcal{K}$ denote the distribution of the known training data and $\mathcal{N}$ denote the distribution of unknown data that is outside of $\mathcal{K}$. Our open set recognition system should determine whether an observation $\mathbf{x}$ is from the known distribution $\mathcal{K}$ or the unknown distribution $\mathcal{N}$. If $\mathbf{x}$ is from $\mathcal{K}$, the classifier should predict a class label $\hat{y} \in \mathcal{K}$, otherwise it should be labelled as unknown/novel.

The deep metric learning model used by our approach \cite{DBLP:journals/corr/MeyerHD17} stores all training set feature embeddings $C$ and computes the Euclidean distance between an example embedding and its set of nearest neighbours for the purpose of classification.
Distance between examples in the metric space can be used as a measure of semantic similarity. We expect that most observed examples from a known class will be located nearby training set embeddings of the same class.
Observed examples that are not located nearby any training set embeddings are novel to the model and are likely from $\mathcal{N}$. Since it is known the metric learning model transfers well to novel classes, we expect the model to be well suited to novelty detection. This assumption is evaluated in Section \ref{novelty_results}.

Our open set classifier and novelty detector predicts a class label $\hat{y}$ for example $\mathbf{x}$ as follows:
\begin{equation} \label{novelty_detection_eq}
\hat{y} = \begin{cases}
    \argmax_{l} Pr(y = l \vert \mathbf{x}), & \text{if $n(\mathbf{x}) \leq \delta$},\\
    \mathrm{unknown/novel}, & \text{if $n(\mathbf{x}) > \delta$},
  \end{cases}
\end{equation}
where $n$ is a novelty function and $\delta$ is a threshold.

We investigate the effectiveness of several simple distance-based novelty measures ($n(\mathbf{x})$) for use in a deep metric space. Distance-based measures are appropriate since we know that the metric learning model transfers well to novel classes.
\subsubsection{\textit{\textbf{Nearest neighbour distance} (NN dist.)}} is the distance between an observed example and its nearest training set embedding. The Euclidean distance between the embeddings $\mathbf{x}$ and $\mathbf{c}_i$ is denoted as $d(\mathbf{x},\mathbf{c}_i)$.
\begin{equation}
n\left(\mathbf{x}\right) = \min_{\mathbf{c}_i \in C} d\left(\mathbf{x},\mathbf{c}_i\right)
\end{equation}
\subsubsection{\textit{\textbf{Maximum class density} (density)}} is found by computing the per class densities of training set embeddings nearby an example, using the Gaussian kernel sum from Equation \ref{gaussian_sum}. For notational convenience in Equation \ref{novelty_detection_eq}, we subtract the class density from one.
\begin{equation}
n\left(\mathbf{x}\right) = 1 - \max_l Pr(y = l \vert \mathbf{x})
\end{equation}
\subsubsection{\textit{\textbf{Entropy}}} is the Shannon entropy of the class density distribution from Equation \ref{gaussian_sum}.
\begin{equation}
n\left(\mathbf{x}\right) = -\sum_l Pr(y = l \vert \mathbf{x}) \log \left( Pr(y = l \vert \mathbf{x}) \right)
\end{equation}

Since the metric learning model's class probability distribution is computed based on class densities, the \textit{density} measure is equivalent to measuring novelty based on the maximum class probability. Density is suggested as a suitable measure for novelty detection in \cite{DBLP:journals/corr/abs-1709-09844} and we adapt the method for our approach, using the shared Gaussian $\sigma$. 

\begin{algorithm}[t]
  \caption{Open set active learning with unlabelled to labelled density ratio (ULDR).
    \label{Al_algorithm}}
  \begin{algorithmic}[1]
	\Require
		\Statex Labelling budget $b$
		\Statex Set of unlabelled feature embeddings (observations) $U$
		\Statex Set of labelled feature embeddings $C$
	\For {$1 \mathellipsis b$}
		\State Initialise $r(i) = 0$, for all $i \in U$
		\ForEach {$i \in U $}
			\State $r(i) = ULDR(i,U,C)$
		\EndFor
		\State $q = \argmax (\mathbf{r})$
		\State Query user for label of $\mathbf{u}_q$
		\State $C$.append$(\mathbf{u}_q)$\Comment{Add to labelled set}
		\State $U$.remove$(q)$\Comment{Remove from unlabelled set}
	\EndFor
	\State Fine-tune deep metric learning model on $C$
	\Statex
	\Function{ULDR}{$i,U,C$}\Comment{Query selection}
      \State \normalsize \label{uldr_line}
      \begin{equation} \label{uldr_eq}
      r = \frac{\sum_{\mathbf{u}_j \in U, j \neq i} \exp{\left(\frac{-\lvert \mathbf{u}_i - \mathbf{u}_j \rvert^2}{2\sigma^2} \right)}}{\sum_{\mathbf{c}_k \in C} \exp{\left(\frac{-\lvert \mathbf{u}_i - \mathbf{c}_k \rvert^2}{2\sigma^2} \right)}}
      \end{equation}
      \State \Return{r}
    \EndFunction
\end{algorithmic}
\end{algorithm}

\begin{table*}[!t]
\caption{Novelty detection Area Under ROC Curve (AUROC), Area Under Precision-Recall Curve (AUPR), F-measure (F1) and open set recognition accuracy (Acc.) on three datasets.}
    \label{novelty_detection_table}
\footnotesize
    \centering
        \begin{tabular}{l@{\hskip 5px}cccccccccccc}\toprule
        & \multicolumn{4}{c}{Cars196} & \multicolumn{4}{c}{Flowers102} & \multicolumn{4}{c}{Birds200} \\
\cmidrule(lr){2-5} \cmidrule(lr){6-9} \cmidrule(lr){10-13}
      Method & AUROC & AUPR & F1 & Acc. & AUROC & AUPR& F1 & Acc.  & AUROC & AUPR& F1 & Acc. \\ \midrule
      Baseline \cite{DBLP:journals/corr/HendrycksG16c} Max Pr.  & 0.8331 & 0.8116 & 0.7832	& 0.7334 & 0.8509 & 0.8051 & 0.8004 & 0.7873 &0.7311 & 0.6933 & 0.7277	& 0.6473\\
      Baseline \cite{DBLP:journals/corr/HendrycksG16c} Entropy & 0.8512 & 0.8374  & 0.7865 & 0.7395  & 0.8559 & 0.8206 & 0.8015 &	0.7907 & 0.7397	& 0.7017 & 0.7280&	0.6422\\
      OpenMax \cite{bendale2016towards} & - & - & 0.8055 & 0.7515 & - & - &  0.7985 & 0.7588 & - & - &  0.7628 & 0.6893 \\
      ODIN \cite{liang2018enhance} Max Pr. & 0.8613 & 0.8443 & 0.8021	& 0.7531 & 0.8712 & 0.8471 & 0.8021	& 0.7531  & 0.7383 & 0.7031 & 0.7271 & 0.6404\\
      ODIN \cite{liang2018enhance} Entropy & 0.8668	& 0.8469 & 0.8089 & 0.7623 & 0.8690 & 0.8447 & 0.8085 & 0.7848 & 0.7400 & 0.7034 & 0.7260 & 0.6389\\
      Ours: DML Density & 0.8901 & 0.8671 & 0.8263 &	0.7878 & 0.9043	& 0.8741 & 0.8442	& 0.8255 & 0.7838	& 0.7475 & 0.7419	& 0.6895 \\
      Ours: DML Entropy & 0.9013	& \textbf{0.8710} & 0.8454	& 0.8033 & \textbf{0.9084}	& 0.8718 & 0.8477	& 0.8299 & \textbf{0.7981} & \textbf{0.7601} & 0.7559	& \textbf{0.7012} \\
      Ours: DML NN Dist. & \textbf{0.9028}	& 0.8706 & \textbf{0.8502}	& \textbf{0.8041} & 0.9078 & \textbf{0.8884} & \textbf{0.8543}	& \textbf{0.8397} & 0.7961	& 0.7489 & \textbf{0.7652} & 0.6840\\ \bottomrule
      \end{tabular}
\end{table*}

\subsection{Active Learning of Novel Classes} \label{active}
When deployed, a deep metric learning model trained on the distribution $\mathcal{K}$, observes new unlabelled data from a mixture of the distributions $\mathcal{K}$ and $\mathcal{N}$. Let $U = [\mathbf{u}_1,\mathellipsis ,\mathbf{u}_\gamma]$ represent the set of feature embeddings from $\gamma$ unlabelled observations, with $\mathbf{u}_i = [u_i^{(1)},\mathellipsis , u_i^{(\beta)}]^T$. In addition to detecting observations belonging to $\mathcal{N}$, our system should select the most informative examples in $U$ for labelling by a user. Obtaining a label is referred to as a \textit{query}. The selected examples should be those that allow the model to learn the most about $\mathcal{N}$, from the fewest queries.
Metric spaces that transfer knowledge to novel examples enable such efficient label querying. Our proposed open set active learning approach is outlined in Algorithm \ref{Al_algorithm}.

We define $b$ as the labelling budget, that is, the number of labels that can be obtained from a user. 
Since $b$ may be significantly smaller than the total number of observations, it is important that examples are selected for querying based on an informativeness measure. This selection process is known as \textit{query selection}.
Once queried, labelled observations are included in the set of Gaussian kernel centres $C$. The network is fine-tuned with the new labelled examples, as well as the original training examples, to ensure that knowledge about previously learned classes is not lost.

Observations should be selected for querying based on two criteria. The first is the novelty of the observation with respect to the labelled training examples. The second is the potential informativeness of a given observation to the set of all unlabelled observations. This criteria means that query selection should favour examples that are both in regions of high unlabelled example density and regions of low labelled example density. In other words, we should select feature embeddings that are far from labelled embeddings and nearby many unlabelled embeddings.

Our proposed approach to query selection, shown in Equation \ref{uldr_eq} (Line \ref{uldr_line} of Algorithm \ref{Al_algorithm}), is the ratio of unlabelled density and labelled density. The value of $\sigma$ is the same as in Equation \ref{gaussian_sum}. The next example selected for querying is that with the largest unlabelled to labelled density ratio (ULDR). When an example is labelled, it is removed from the set of unlabelled observations $U$ and included in the set of labelled examples $C$, which includes the original training data. The ULDR query selection method ensures examples that are both far from labelled examples and in regions of high unlabelled density are favoured. This is important because a lone unlabelled observation is less informative than one with high unlabelled density. It is likely that a cluster of novel observations indicates the presence of a class that is outside of $\mathcal{K}$, but common in $\mathcal{N}$. Such observations should be the first to be queried. 

As discussed in Section \ref{dml_section}, fast approximate nearest neighbour search and period asynchronous updates of the training embeddings can be utilised to make query selection and network fine-tuning scalable to large numbers of classes and training examples. Nearest neighbours are computed to classify an observation and can be used to consider only a local neighbourhood of training examples for the ULDR computation. These details are discussed in depth in \cite{DBLP:journals/corr/MeyerHD17}.

\section{Experiments} \label{results}

\subsection{Experimental Set-up} \label{setup}
We evaluate our deep metric learning approaches to open set recognition and active learning of novel classes on three datasets: Stanford Cars196 \cite{krause20133d}, Oxford Flowers102 \cite{Nilsback08} and CUB Birds200 2011 \cite{WelinderEtal2010}. For each dataset, the first half of classes, that is, the first 98, 51 and 100 classes respectively, are taken as known classes (from $\mathcal{K}$). The remaining half are taken as novel classes (from $\mathcal{N}$). The datasets are split into training, observed and test sets. The training sets contain only known class images, while the observed and test sets contain an equal number of known and novel class images.

A VGG16 \cite{Simonyan14c} architecture is used for all experiments, as we find this network configuration performs well for transfer learning tasks. The second fully connected layer, FC7, is taken as the embedding layer. This produces a 4096 dimension feature embedding for a given input image, and therefore, a 4096 dimension metric space. The network is trained on the training set of known classes, following the methodology described in \cite{DBLP:journals/corr/MeyerHD17}. 
Training data is augmented using random cropping and horizontal mirroring. A learning rate of 0.00001, weight decay of 0.0005, momentum of 0.9 and shared Gaussian kernel $\sigma$ of 91, 75 and 103 is used for Cars196, Flowers102 and Birds200, respectively. All hyperparameters are selected as described in \cite{DBLP:journals/corr/MeyerHD17}.

\begin{figure*}[!t]
\begin{tabular}{ccc}
\includegraphics[width=5.5cm]{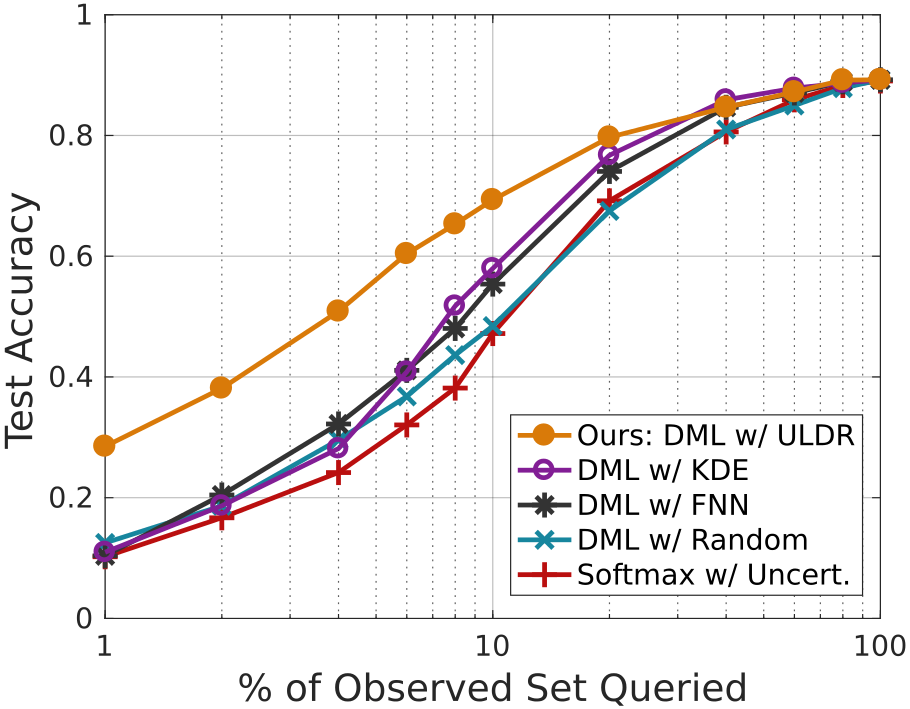}&
\includegraphics[width=5.5cm]{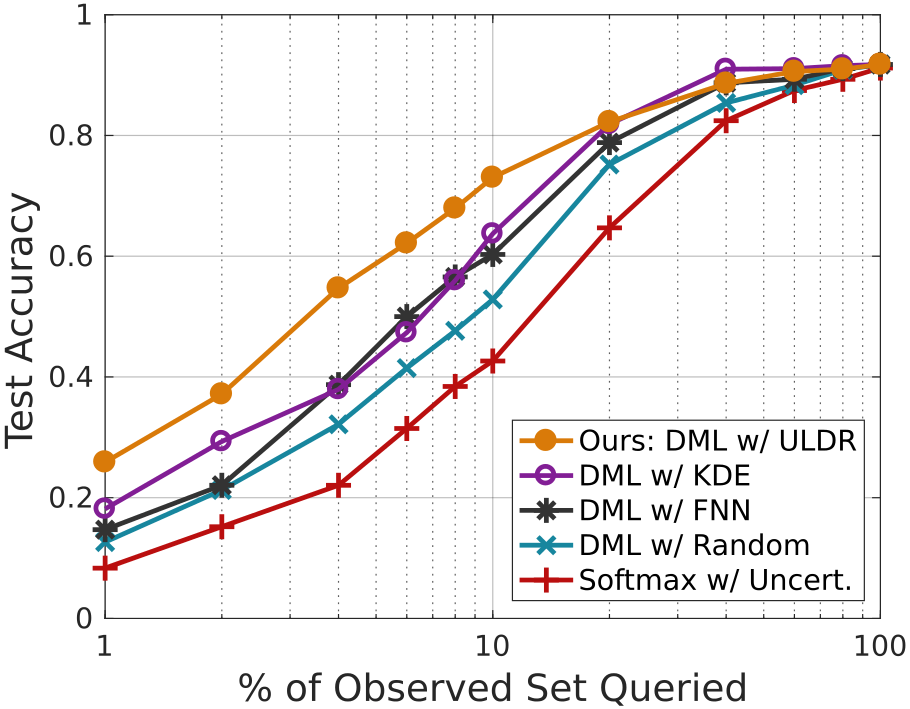}&
\includegraphics[width=5.5cm]{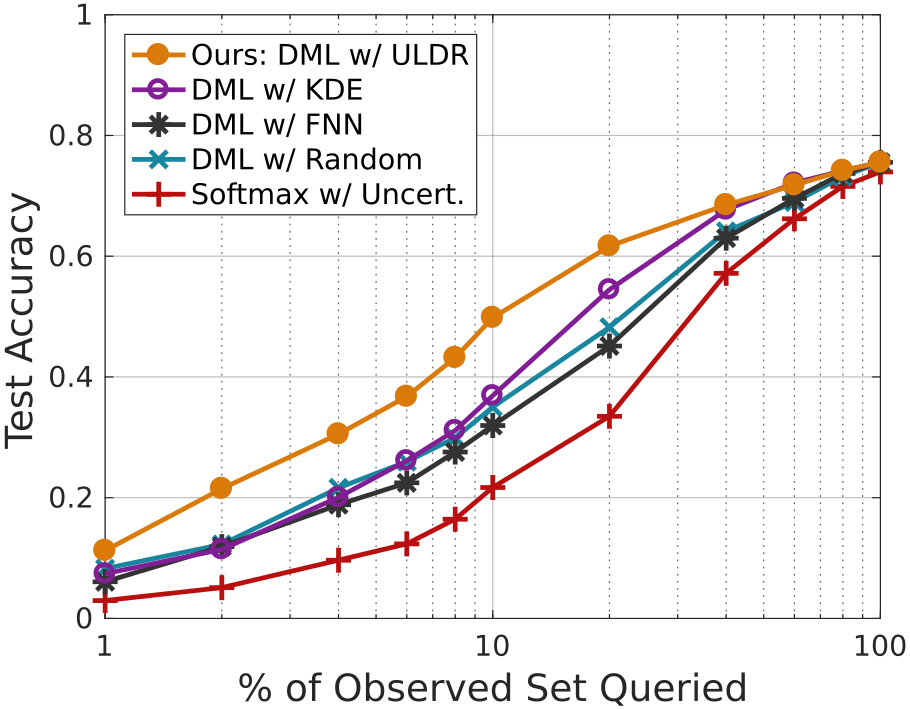}\\
\hspace*{0.5cm}\scriptsize(a) Cars196, novel only.&\hspace*{0.5cm}\scriptsize(b) Flowers102, novel only.&\hspace*{0.5cm}\scriptsize(c) Birds200, novel only.\vspace{0.25cm}\\
\includegraphics[width=5.5cm]{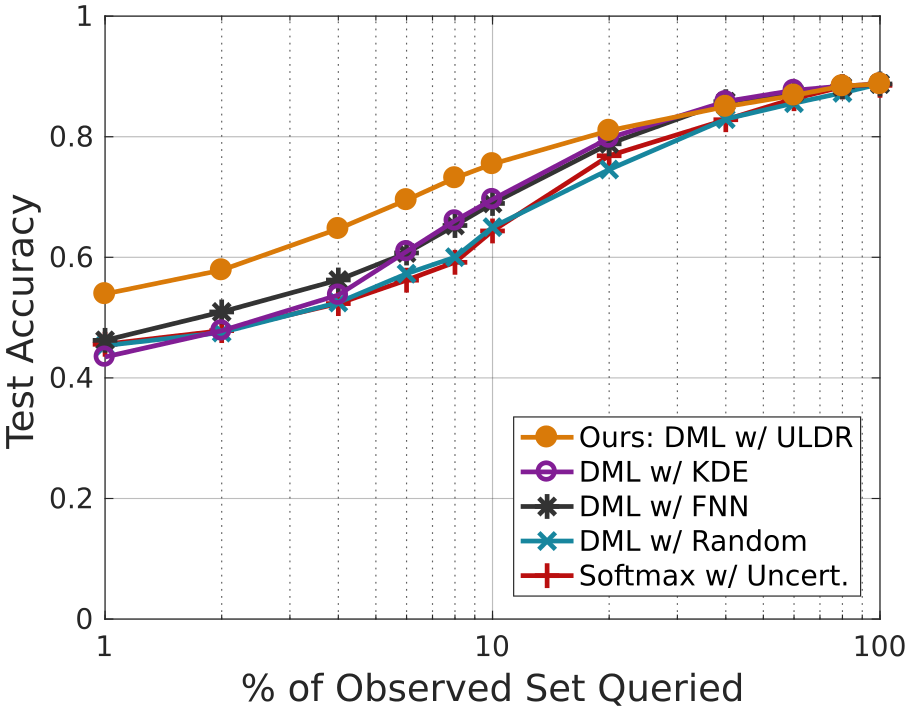}&
\includegraphics[width=5.5cm]{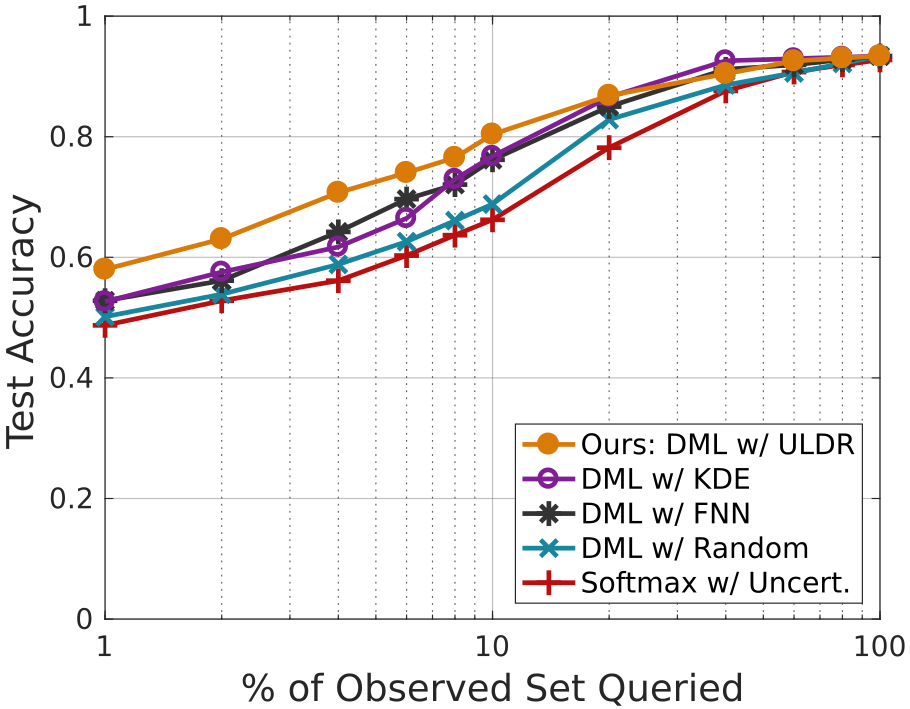}&
\includegraphics[width=5.5cm]{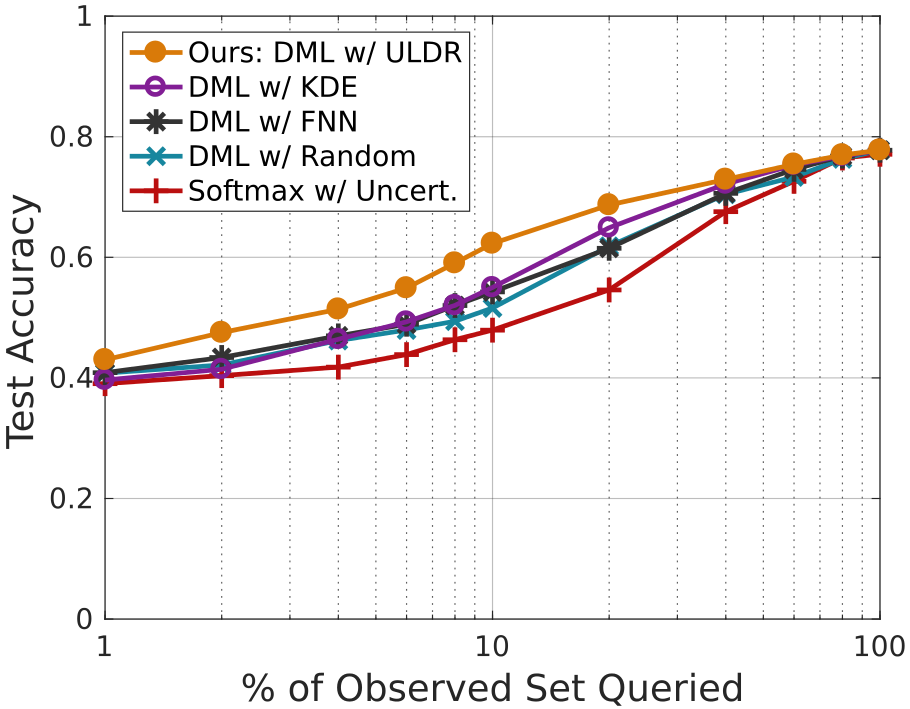}\\
\hspace*{0.5cm}\scriptsize(d) Cars196, novel and known.&\hspace*{0.5cm}\scriptsize(e) Flowers102, novel and known.&\hspace*{0.5cm}\scriptsize(f) Birds200, novel and known.\normalsize \\
\end{tabular}
\caption{Active learning: Plots of test set classification accuracy of novel classes only and combined novel and known classes at various labelling budgets. The labelling budget is represented as a percentage of the total observed set.}
\label{active_learning_plot}
\end{figure*}

\subsection{Novelty Detection and Open Set Recognition Results} \label{novelty_results}

The network is first trained on the training set of known classes. The observed set, containing examples from both known and novel classes, is then used for the evaluation. We compare a deep metric learning (\textit{DML}) approach with differing novelty measures (\textit{NN dist.}, \textit{density} and \textit{entropy}, as discussed in Section \ref{anomaly_detection}) with the following baselines and state-of-the-art novelty detectors and open set classifiers:
\begin{itemize}
	\item \textit{Baseline \cite{DBLP:journals/corr/HendrycksG16c} Max Pr.}: A baseline softmax uncertainty novelty detector with maximum class probability thresholding.
	\item \textit{Baseline \cite{DBLP:journals/corr/HendrycksG16c} Entropy}: A baseline softmax uncertainty novelty detector with Shannon entropy thresholding.
	\item \textit{OpenMax \cite{bendale2016towards}}: Open set version of softmax classification.
	\item \textit{ODIN \cite{liang2018enhance} Max Pr.}: Out-of-distribution detector with maximum class probability thresholding.
	\item \textit{ODIN \cite{liang2018enhance} Entropy.}: Out-of-distribution detector with Shannon entropy thresholding.
\end{itemize}

We evaluate novelty detection with the Area Under ROC Curve (AUROC) and Area Under Precision-Recall Curve (AUPR) measures, avoiding threshold selection. We further evaluate with a fixed threshold, analysing the novelty detection F-measure, i.e. $F1 = 2 \frac{(precision \cdot recall)}{precision+recall}$, and open set recognition accuracy ($Acc.$), which is the standard classification accuracy with a single unknown/novel superclass for all observations from $\mathcal{N}$. Both our approach and \cite{DBLP:journals/corr/HendrycksG16c} have only one tunable parameter (the threshold $\delta$), while \cite{bendale2016towards} and \cite{liang2018enhance} each have three. A withheld set of images is used to tune parameters such that the withheld set F-measure is maximised, as suggested in \cite{bendale2016towards}. Note that no parameter tuning is needed for the AUROC and AUPR measures for our approach or the softmax baseline \cite{DBLP:journals/corr/HendrycksG16c}. Since OpenMax \cite{bendale2016towards} explicitly includes a novel pseudo-class probability, AUROC and AUPR measures cannot be computed. As such, we report only the F-measure and open set accuracy for this approach. Results are shown in Table \ref{novelty_detection_table}. Our approach outperforms the compared methods on all evaluation measures and datasets, in most cases by a significant margin.

\begin{figure*}[!t]
\begin{tabular}{cccc} 
\includegraphics[width=4cm]{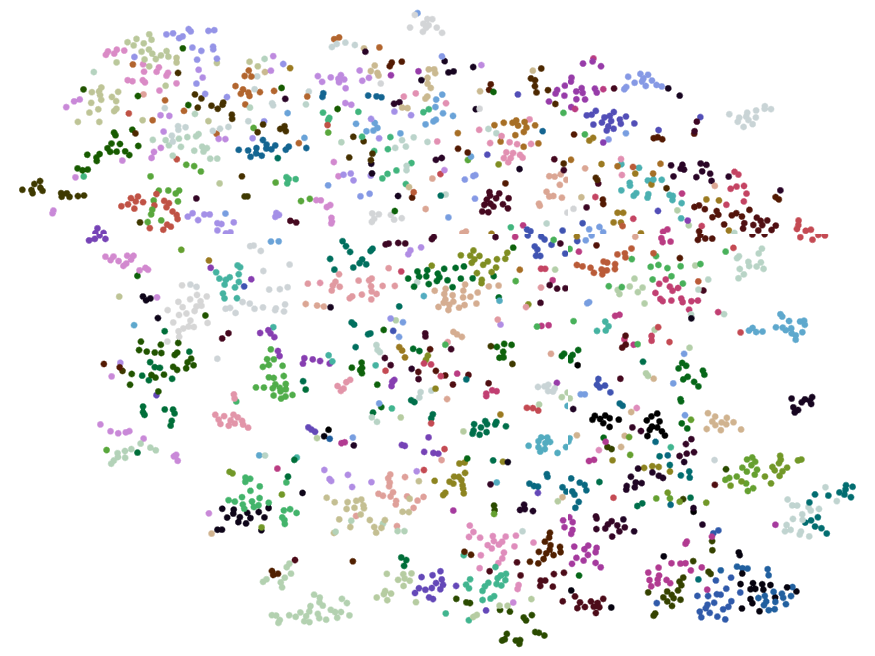}&
\includegraphics[width=4cm]{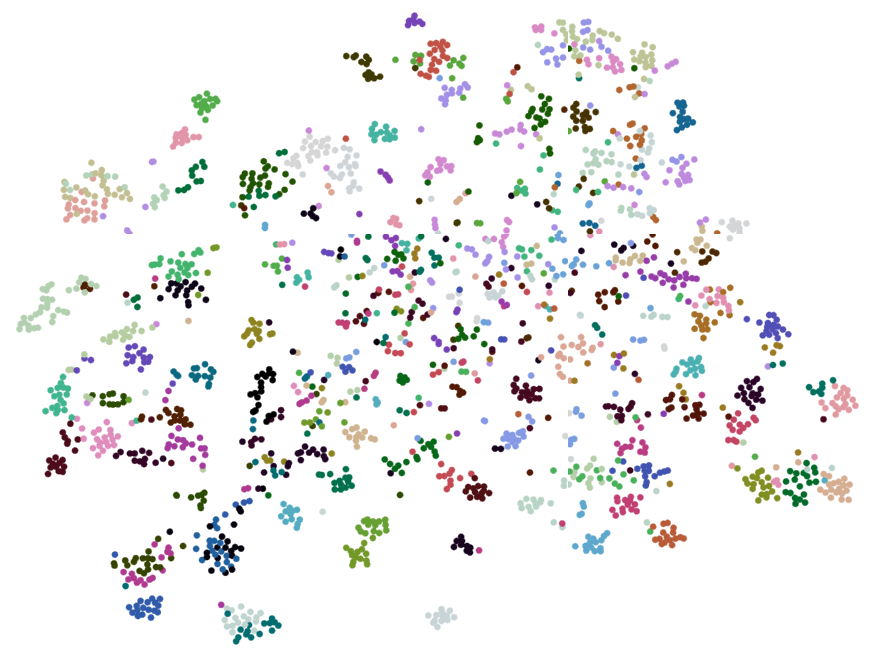}&
\includegraphics[width=4cm]{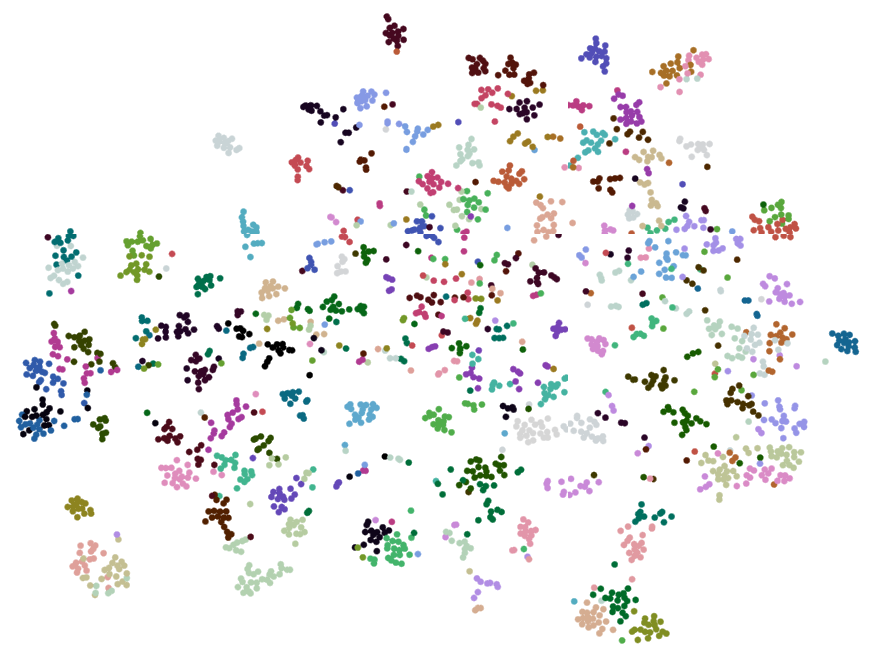}&
\includegraphics[width=4cm]{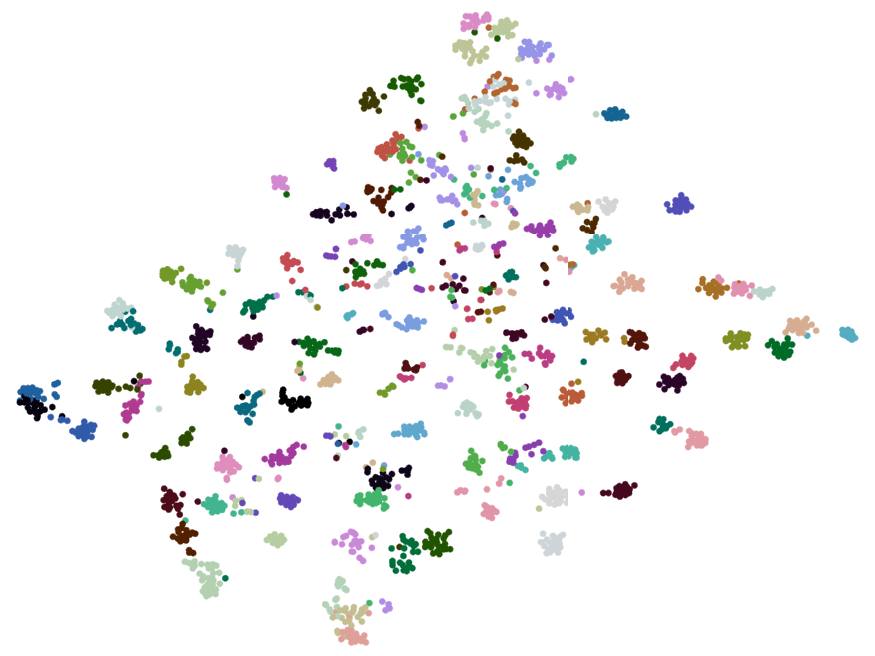}\\
\small(a) No fine-tune on observed set. & \small(b) Pseudo-label fine-tune.\vspace{0.5cm} & \small(c) AL with b = 10\%. & \small(d) AL with b = 100\%. \normalsize
\end{tabular}
\caption{Visualisations of the metric space of novel class test examples using the t-SNE algorithm \cite{maaten2008visualizing} on Cars196. Colour represents the class of examples. The initial metric space before fine-tuning on the observed set is shown in (a), while (b) is the metric space after fine-tuning using pseudo-labels only (Section \ref{pseudo_labels_sec}). Metric spaces after active learning (\textit{AL}) are shown in (c) and (d), with the labelling budget $b$ as a percentage of the observed set labelled. A budget of 100\% is the upper bound on performance. Novel examples are already quite well clustered before any fine-tuning on novel classes, as seen in (a). This demonstrates the ability of deep metric spaces to transfer well to novel classes. This is the property that motivates our approach.}
\label{unsupervised_plot}
\end{figure*}

\subsection{Open Set Active Learning Results} \label{active_results}
The network is first trained on the training set of known classes. Query selection is then carried out on the observed set, which contains examples from both known and novel classes. The network is then fine-tuned with the selected observed set examples, together with the original training set examples. In our experiments, labels are provided to the model automatically in response to a query. This simulates the process of a human user providing labels to a robot. We evaluate on the test set, containing unseen examples from both the original known class set and the novel class set. Classification accuracy on both the novel classes only and the combined novel and previously known classes is reported. Note that individual novel classes are used for the accuracy calculation in this section, not a single superclass, as in Section \ref{novelty_results}. The following approaches are compared:
\begin{itemize}
  \item \textit{Softmax w/ Uncert.}: A conventional softmax approach with a typical query selection method based on classifier uncertainty. The observation with the largest Shannon entropy is queried.
  \item \textit{DML w/ Random}: Deep metric learning approach with random query selection.
  \item \textit{DML w/ FNN \cite{5688457}}: Deep metric learning approach with furthest nearest neighbour (FNN) query selection \cite{5688457}. The observation with the largest distance to its nearest labelled example in the metric space is queried.
  \item \textit{DML w/ KDE}: Deep metric learning approach with kernel density estimation (KDE) query selection. The observation with the smallest maximum class probability (density) from Equation \ref{gaussian_sum} is queried.
  \item \textit{Ours: DML w/ ULDR}: Deep metric learning with our unlabelled to labelled density ratio (ULDR) query selection. The observation with the largest ULDR is queried.
\end{itemize}

Results are shown in Figure \ref{active_learning_plot}. Note that the metric learning model from \cite{DBLP:journals/corr/MeyerHD17} is used for all \textit{DML} methods. Our experiments aim show two important points: that deep metric learning is better suited to open set active learning than softmax-based networks, and that our proposed ULDR approach to query selection is efficient and effective. The softmax-based approach is outperformed by even random query selection with a deep metric learning model. Our proposed ULDR query selection method significantly outperforms the compared approaches for small labelling budgets. This shows how our method would allow a robot to efficiently query a user for labels, minimising the number of queries needed to achieve a given performance and therefore minimising the required human effort. For labelling budgets of less than 10\%, our approach outperforms the nearest compared method by an average of 16.3\% on Cars196.

A t-SNE visualisation \cite{maaten2008visualizing} of the novel class test set metric space is shown in Figure \ref{unsupervised_plot}. Novel test classes are already quite well clustered before any learning has taken place with novel examples. This shows the transfer learning capabilities of deep metric learning that motivate our approach.

\begin{table}[t]
\caption{Unsupervised pseudo-label approach compared to no novel class fine-tuning and active learning of novel classes.}
\label{pseudo_label_table}
\begin{center}
\begin{tabular}{lp{0.7cm}p{0.7cm}p{0.7cm}p{0.7cm}p{0.9cm}}
\toprule 
 & Novel R@1 & Novel R@2 & Novel R@4 & Novel R@8 & Known Acc.  \\
\midrule
Initial  & .6893 & .7911 & .8680 & .9161 & .8322 \\
\midrule
Pseudo-labels  & .7504 & .8316 & .8904 & .9306 & .8368 \\
\midrule
AL, b=10\% & .7742 & .8447 & .8978 & .9360 &   .8387 \\
AL, b=100\% & .8680 & .9101 & .9484 & .9687 & .8901 \\
\bottomrule
\end{tabular}
\end{center}
\end{table}

\subsection{Improving Novel Class Representation with Zero Labelling Budget} \label{pseudo_labels_sec}
We further investigate whether a model can improve its representation of observed novel classes with a labelling budget of zero.
We use spatial relationships in the metric space to generate pseudo-labels for observed examples. In other words, the knowledge that deep metric spaces transfer well to novel classes is used to generate a training signal. We use $k$-means \cite{lloyd1982least}, with $k$-means++ initialisation \cite{Arthur:2007:KAC:1283383.1283494}, to obtain pseudo-labels for each observed example. The network is fine-tuned using the observed examples with pseudo-labels together with the original training examples. The value of $k$ is selected such that the Silhouette Score \cite{rousseeuw1987silhouettes} is maximised, indicating that the cluster assignments are tight. A $k$ value of 240 is used for the Cars196 dataset. Since the true labels of the observed set are not known in this case, we evaluate how well the network has learned to represent the novel set of classes using a recall measure on the test set examples. Recall@m (R@m) is the fraction of test examples that have the same true class label as at least one of their $m$ nearest neighbours in the metric space. Experiments are run several times and the results are averaged.

Table \ref{pseudo_label_table} shows the Recall@m of the novel examples in the Cars196 test set. Note that the test metric space contains examples from both known and novel classes. The classification accuracy of the test set examples from known classes (\textit{Known Acc.}) is also shown. The pseudo-label approach is compared to a model that has not been fine-tuned on the observed set (\textit{initial}). This is the lower bound on performance. Active learning (\textit{AL}) results are also included, with labelling budgets $b$ of 10\% and 100\%. The 100\% labelling budget is the upper bound on performance, as the entire observed set is labelled. Figure \ref{unsupervised_plot} shows a t-SNE visualisation \cite{maaten2008visualizing} of the novel test set metric space. Compared to the initial metric space, novel classes are better clustered. Interestingly, these results indicate that although no true labels are available for the observed set, there is merit in allowing observed examples to be pushed into a better region of the metric space. We do not expect known class accuracy to improve with this method, but importantly, it does not deteriorate (see final column of Table \ref{pseudo_label_table}).

\section{Conclusion}
In this paper, the suitability of deep metric learning to open-set robotic vision problems was investigated.
We showed how a deep metric learning classification model is well suited to novelty detection and open set recognition. A novel approach to the active learning of previously unknown classes was also proposed.
At small labelling budgets, our approach significantly outperforms comparable methods. This would allow a robotic vision system to efficiently and effectively extend its understanding of the environment beyond the original training distribution.

\bibliographystyle{IEEEtran}

\end{document}